\documentclass[letterpaper, 10 pt, conference]{ieeeconf}  

\IEEEoverridecommandlockouts                              

\let\labelindent\relax

\usepackage{graphics} 
\graphicspath{ {./Figures/} }
\usepackage{epsfig} 
\usepackage{mathptmx} 
\usepackage{times} 
\usepackage{amsmath} 
\usepackage{amssymb}  
\usepackage{mathtools}
\usepackage{algorithm}
\usepackage{algpseudocode}
\usepackage{xcolor}
\usepackage{enumitem}

\makeatletter
\newenvironment{breakablealgorithm}
  {
   \begin{center}
     \refstepcounter{algorithm}
     \hrule height.8pt depth0pt \kern2pt
     \renewcommand{\caption}[2][\relax]{
{\raggedright\textbf{\fname@algorithm~\thealgorithm} ##2\par}%
       \ifx\relax##1\relax 
         \addcontentsline{loa}{algorithm}{\protect\numberline{\thealgorithm}##2}%
       \else 
         \addcontentsline{loa}{algorithm}{\protect\numberline{\thealgorithm}##1}%
       \fi
       \kern2pt\hrule\kern2pt
     }
  }{
     \kern2pt\hrule\relax
   \end{center}
  }
\makeatother

\title{\LARGE \bf
Collision Avoidance of 3-Dimensional Objects in Dynamic Environments
}

\author{ Kashish Dhal$^{1}$ Abhishek Kashyap$^{1}$  Animesh Chakravarthy$^{1}$    
\thanks{$^{1}$Kashish Dhal, Abhishek Kashyap and Animesh Chakravarthy are with the Department of Mechanical and Aerospace Engineering, University of Texas at Arlington, TX, USA{\tt\small kashish.dhal@mavs.uta.edu, abhishek.kashyap@mavs.uta.edu, animesh.chakravarthy@uta.edu}}%
}

\begin{document}
\maketitle
\thispagestyle{empty}
\pagestyle{empty}

\begin{abstract}
Achieving collision avoidance between moving objects is an important objective while determining  robot trajectories.  In performing collision avoidance maneuvers, the relative shapes of the objects play an important role.  The literature largely models the shapes of the objects as spheres, and this can make the avoidance maneuvers very conservative, especially when the objects are of elongated shape and/or non-convex.  In this paper, we model the shapes of the objects using suitable combinations of ellipsoids and one-sheeted/two-sheeted hyperboloids, and employ a collision cone approach to achieve collision avoidance.  We present a method to construct the 3-D collision cone, and present simulation results demonstrating the working of the collision avoidance laws.
\end{abstract}

\section{INTRODUCTION}

 An important component of robot path planning is the collision avoidance problem, that is, determining a safe trajectory of a robotic vehicle so that it circumvents various obstacles in its path.  When the robot and obstacles are operating in close proximity, their relative shapes can play an important role in the determination of collision avoidance trajectories.  One common practice is to use polygonal approximations as bounding boxes for the shapes of the robots and obstacles.  However, the polygonal approximation can lead to increased computational complexity (measured in terms of obstacle complexity, or the amount of information used to store 
a computer model of the obstacle, where obstacle complexity is measured in terms of the number of obstacle edges \cite{c6}).  To overcome this, a common practice then is to use spherical approximations for the robots and obstacles, because of the analytical convenience of such approximations, along with the reduced information required to store a computational model of the obstacle.  The obstacle avoidance conditions are then computed for the sphere as a whole.  
\begin{figure}
\centering
\includegraphics[width=0.35\textwidth]{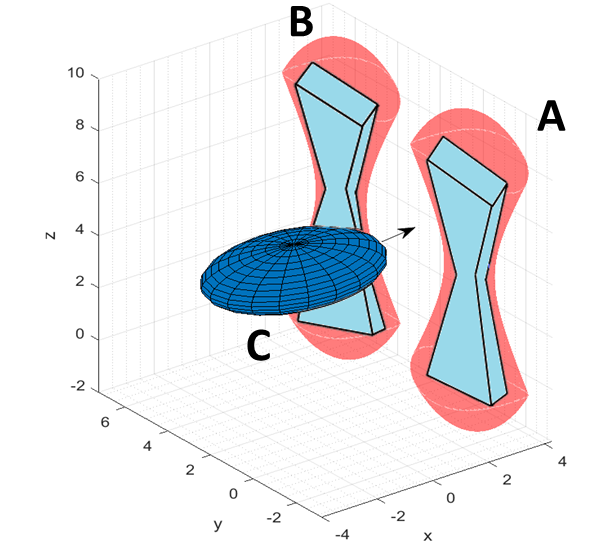}
\caption{Elongated and/or non-convex agents in a confined space}
\vspace{-30pt}
\label{fig:two_pillars}
\end{figure}

These approximations become overly conservative in cases when an object is
more elongated along one dimension compared to others. For instance, refer Fig \ref{fig:two_pillars} comprising three moving elongated agents.  If we approximate each of $A$ and $B$ with a sphere, then these two spheres will intersect and as a consequence,  $C$ will deem there is no path for it to go between $A$ and $B$, even though such a path exists.  To overcome this, one can model the shapes of $A$ and $B$ using multiple smaller spheres, but this can increase the computational load.  In such cases, ellipsoids have been used to serve 
as better approximations for the object shapes \cite{harshat2019Ellipsoidal}, \cite{c9}.  For non-convex objects such as star-shaped objects ~\cite{harshat2020Star} however, even ellipsoidal approximations can become over conservative, because such approximations reduce the amount of available free space within which the robot trajectories can lie.  In such cases, one can take recourse to non-convex bounding approximations involving a combination of ellipsoids and hyperboloids.       

This paper employs a collision cone based approach to determine collision avoidance laws for moving objects having elongated, non-convex shapes.  The collision cone approach, originally introduced in \cite{c2}, has some similarities with the velocity obstacle approach \cite{Fio} in that both approaches determine the set of velocities of the robots that will place them on a collision course with one or more obstacles.  However while the velocity obstacle approach, and its many extensions \cite{manocha}, has been largely restricted to circles/spheres, the fact that the collision cone approach has its roots in missile guidance, enables it to determine closed form collision conditions for a larger class of object shapes \cite{c2},\cite{c3},\cite{c4}.  The great benefit of obtaining analytical expressions of collision conditions is that these then serve as a basis for designing collision avoidance laws.  The collision cone approach of \cite{c2} has been extensively employed in the literature (See for example, \cite{Ferrara1},\cite{Watanabe2},\cite{vision},\cite{lalish2},\cite{biological},\cite{zuo2020model}).  

When a pair of agents have different shapes, a way to superpose the shape of one agent onto the other is to perform a Minkowski sum operation.  However, this may be computationally expensive.  In \cite{ownCDCpaper}, the authors computed the collision cone between moving quadric surfaces on a plane without taking recourse to computing the Minkowski sum.  In this paper, we consider a larger class of objects moving in 3-D environments and compute the 3-D collision cone between pairs of (differently shaped) 3-D objects, without computing the Minkowski sum.  
    
\section{Equations of 3-D Quadric Surfaces} \label{eqn}
In this section, we present a discussion of the 3-D shapes that occur as a consequence of combining different types of quadrics.      
The equation of a general 3-D quadric is:
\begin{equation}\label{eqn:ellipsoid1}
\begin{split}
&a_{xx}x^2+a_{yy}^2+a_{zz}z^2+2a_{xy}xy+2a_{yz}yz+2a_{xz}xz \\
&+2b_{x}x+2b_{y}y+2b_{z}z+c_1 = 0
\end{split}
\end{equation}

Eqn (\ref{eqn:ellipsoid1}) can equivalently be written in matrix form as follows:
\begin{eqnarray} \label{eqn:ellipsoid2}
\underbrace{ \begin{bmatrix}
        x & y & z & 1
        \end{bmatrix}}_{\bf{x}^T}
   \underbrace{ \begin{bmatrix}
       a_{xx}&a_{xy}&a_x&b_{x}\\
       a_{xy}&a_{yy}&a_{yz}&b_{y}\\
       a_{xz}&a_{yz}&a_{zz}&b{z}\\
       b_{x}&b_{y}&b_z&c_1
        \end{bmatrix} }_{\bf{Q}}
      \underbrace{   \begin{bmatrix}
        x\\
        y\\
        z\\
        1
        \end{bmatrix}}_{\bf{x}}&=&0 
\end{eqnarray}
When $det(\mathbf{Q})<0$, (\ref{eqn:ellipsoid1}) represents an ellipsoid or a one-sheeted hyperboloid, and when $det(\mathbf{Q})>0$, it represents a two-sheeted hyperboloid.    We refer to the matrices corresponding to an ellipsoid, one-sheeted hyperboloid and two-sheeted hyperboloid as $\mathbf{Q_e}$, $\mathbf{Q_{h1}}$ and $\mathbf{Q_{h2}}$, respectively. Please also note that in this paper, vectors are represented in lowercase boldface, and matrices in uppercase boldface.

We now proceed towards determination of equations of surfaces comprised of different combinations of the above quadric surfaces.
\begin{figure}
\centering
\includegraphics[width=0.35\textwidth]{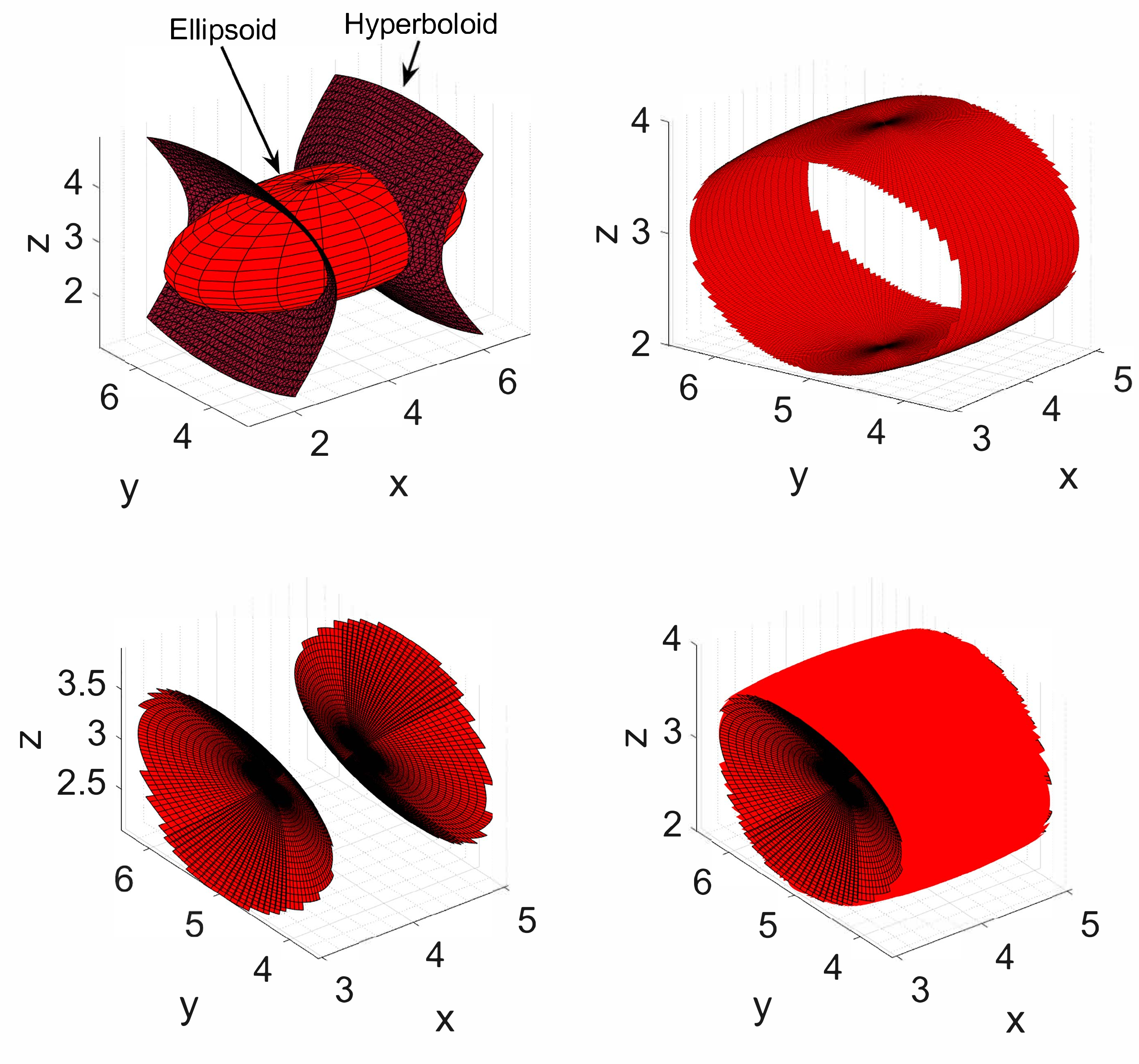}
\includegraphics[width=0.25\textwidth]{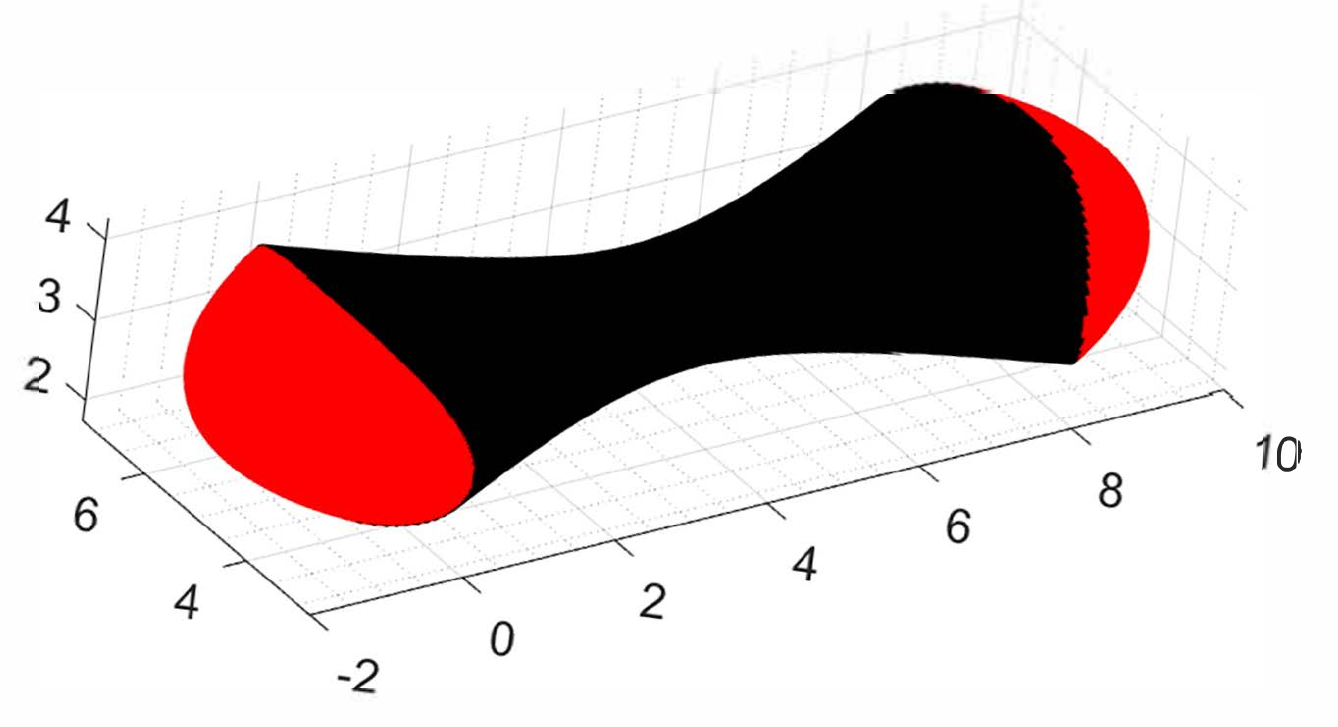}
\caption{a) Ellipsoid and Hyperboloid b) Ellipsoid delimited by Hyperboloid c) Hyperboloid delimited by Ellipsoid d) Biconcave Ellipsoid e) Biconvex Hyperboloid} 
\vspace{-15pt}
\label{fig:quadric}
\end{figure}
%
We define the interior of a quadric as the region which includes the center of the quadric and the exterior as the complement of the interior.  Accordingly, the regions $\{x:\mathbf{x}^T \mathbf{Q}_{e} \mathbf{x} < 0$\} and $\{x:\mathbf{x}^T \mathbf{Q}_{h1} \mathbf{x} < 0\}$ represent,  respectively, the interiors of the ellipsoid corresponding to $Q_e$, and the one-sheeted hyperboloid corresponding to $Q_{h1}$.  On the other hand, the region $\{x:\mathbf{x}^T \mathbf{Q}_{h2} \mathbf{x} < 0$\} represents the exterior of the two-sheeted hyperboloid corresponding to $Q_{h2}$.  
We now use these properties to construct surfaces that comprise combinations of two intersecting quadrics.  In constructing these combinations, we employ the phrase ``delimited'', which means ``having fixed boundaries or limits''.

Consider an intersecting ellipsoid and two-sheeted hyperboloid, as shown in Fig \ref{fig:quadric}a.    Then, we define the surface of an Ellipsoid Delimited by a Hyperboloid (EDH) as follows: 
\begin{eqnarray} \label{eqn:EDH}
   \mathbf{x}^T \mathbf{Q_e} \mathbf{x} = 0,\ \text{subject to:} 
    \bf{x}^T \bf{Q}_{h2} \bf{x} > 0
\end{eqnarray}
The above equation states that the surface of the EDH comprises those points on the surface of the ellipsoid that are not present inside the two-sheeted hyperboloid.  The surface of an EDH is shown in Fig \ref{fig:quadric}b.

We next define the surface of a two-sheeted Hyperboloid Delimited by an Ellipsoid (HDE) as follows:
\begin{eqnarray}  \label{eqn:HDE}
   \mathbf{x}^T \mathbf{Q_{h2}} \mathbf{x} = 0,\ \text{subject to:} 
    \bf{x}^T \bf{Q}_e \bf{x} < 0
\end{eqnarray}
The above equation states that the surface of a HDE comprises those points on the surface of a two-sheeted hyperboloid that are present inside the ellipsoid.  Such a HDE is shown in Fig \ref{fig:quadric}c.

We can combine (\ref{eqn:EDH}) and (\ref{eqn:HDE}) to determine a surface composed of an EDH and a HDE.  This is shown in Fig \ref{fig:quadric}d, and is mathematically represented as:
\begin{equation} \label{eqn:biconcave}
    K_1 \mathbf{x}^T \mathbf{Q_e} \mathbf{x} + K_2 \mathbf{x}^T  \mathbf{Q_{h2}} \mathbf{x} = 0 \textrm{, where}
\end{equation}
$$
K_1 = \begin{cases}1\ if\ \{x: \mathbf{x}^T \mathbf{Q_{h2}} \mathbf{x} > 0 \} \\
 0\ \textrm{otherwise} \end{cases}, 
K_2 = \begin{cases}1\ if\ \{x: \mathbf{x}^T \mathbf{Q_e} \mathbf{x} < 0\} \\
 0\ \textrm{otherwise} \end{cases}
$$
With some abuse of terminology, we refer to the above as a biconcave ellipsoid.  
Finally, we combine a one-sheeted hyperboloid and an ellipsoid.  Its mathematical representation is as follows:
\begin{equation} \label{eqn:biconcave1}
    K_1 \mathbf{x}^T \mathbf{Q_e} \mathbf{x} + K_2 \mathbf{x}^T  \mathbf{Q_{h1}} \mathbf{x} = 0 \textrm{, where}
\end{equation}
$$
K_1 = \begin{cases}1\ if\ \{x: \mathbf{x}^T \mathbf{Q_{h1}} \mathbf{x} < 0 \} \\
 0\ \textrm{otherwise} \end{cases}, 
K_2 = \begin{cases}1\ if\ \{x: \mathbf{x}^T \mathbf{Q_e} \mathbf{x} < 0\} \\
 0\ \textrm{otherwise} \end{cases}
$$
This is shown in Fig \ref{fig:quadric}e.  With some abuse of terminology, we refer to this as a biconvex hyperboloid.  We note that by  combining an ellipsoid with multiple hyperboloids at different orientations, one can also approximate star-shaped objects.   


\section{3D engagement geometry}

 \begin{figure}
\centering
\includegraphics[width=0.33\textwidth]{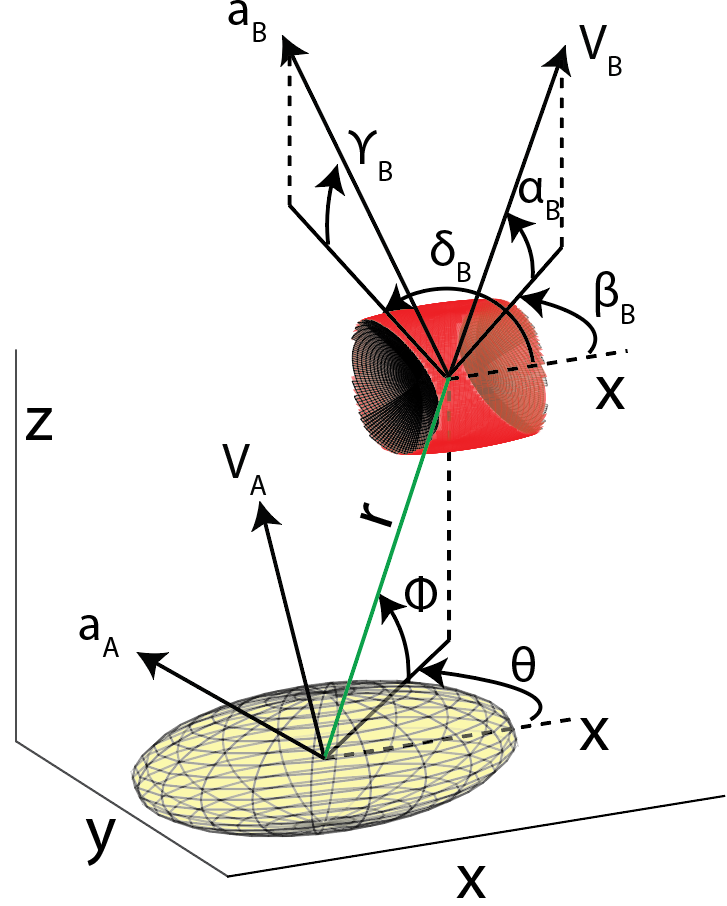}
\caption{Engagement Geometry between two objects}
\vspace{-20pt}
\label{fig:enagement_geometry}
\end{figure}
Fig \ref{fig:enagement_geometry} shows the engagement geometry between two objects $A$ and $B$.  While the figure shows $A$ and $B$ to be an ellipsoid and a biconcave ellipsoid, they could in principle be any pair of objects discussed in Section \ref{eqn}.  $A$ and $B$ are moving with speeds $V_A$ and $V_B$, respectively, at heading angle pairs of $(\beta_A,\alpha_A)$, and $(\beta_B,\alpha_B)$ respectively. Here, $\beta_A$ and $\alpha_A$ represent, respectively, the azimuth and elevation angles of the velocity vector of $A$, and a corresponding definition holds for $\beta_B$ and $\alpha_B$. $r$ represents the distance between the centers of $A$ and $B$, and $(\theta,\phi)$ represents the azimuth-elevation angle pair of line joining the center of $A$ and $B$. The control input of $A$ is its  lateral accelerations $a_{A}$, which acts normal to the velocity vector of $A$ at an azimuth-elevation angle pair of ($\delta_A,\gamma_A$).  A corresponding definition holds for the lateral acceleration $a_{B}$ of $B$.  $V_r,V_\theta,V_\phi$ represent the mutually orthogonal components of the relative velocity of $B$ with respect to $A$, where $V_r$ acts along the line joining the centers of $A$ and $B$.  The kinematics governing the engagement geometry are characterized by the following: 
\begin{equation}\label{eq:state_kinematics}
\begin{split}
    &\dot{r} = V_r, \,\,\dot{\theta} = V_\theta/(r\cos\phi),\,\, \dot{\phi} = V_\phi/r\\
    &\dot{V}_\theta = (-V_\theta V_r+V_\theta V_\phi\tan\phi)/r-(\cos\gamma_A\sin(\delta_A-\theta))a_A\\
    &\quad\quad +(\cos\gamma_B\sin(\delta_B-\theta))a_B\\
    &\dot{V}_\phi = (-{V_{\theta}}{V_{r}} - V_{\theta}^2 \tan{\phi} )/{r}+(\cos{\gamma_A}\sin{\phi}\cos{(\delta_A - \theta)}\\
    &- \sin{\gamma_A}\cos{\phi})a_A-(\cos{\gamma_B}\sin{\phi}\cos{(\delta_B - \theta)} - \sin{\gamma_B}\cos{\phi})a_B\\
    &\dot{V}_r = (V_{\theta}^2 + V_{\phi}^2)/{r}-(\cos{\gamma_A}\cos{\phi}\cos{(\delta_A - \theta)} + \sin{\gamma_A}\sin{\phi})a_A\\
    &+(\cos{\gamma_B}\cos{\phi}\cos{(\delta_B - \theta)} + \sin{\gamma_B}\sin{\phi})a_B
\end{split}
\end{equation}

\vspace{-15pt}
\section{Collision Cone Computation}\label{sec:Collcone}

\subsection{2D Collision Cone for arbitrarily shaped objects} \label{background}

\begin{figure}
\centering
\includegraphics[width=0.35\textwidth]{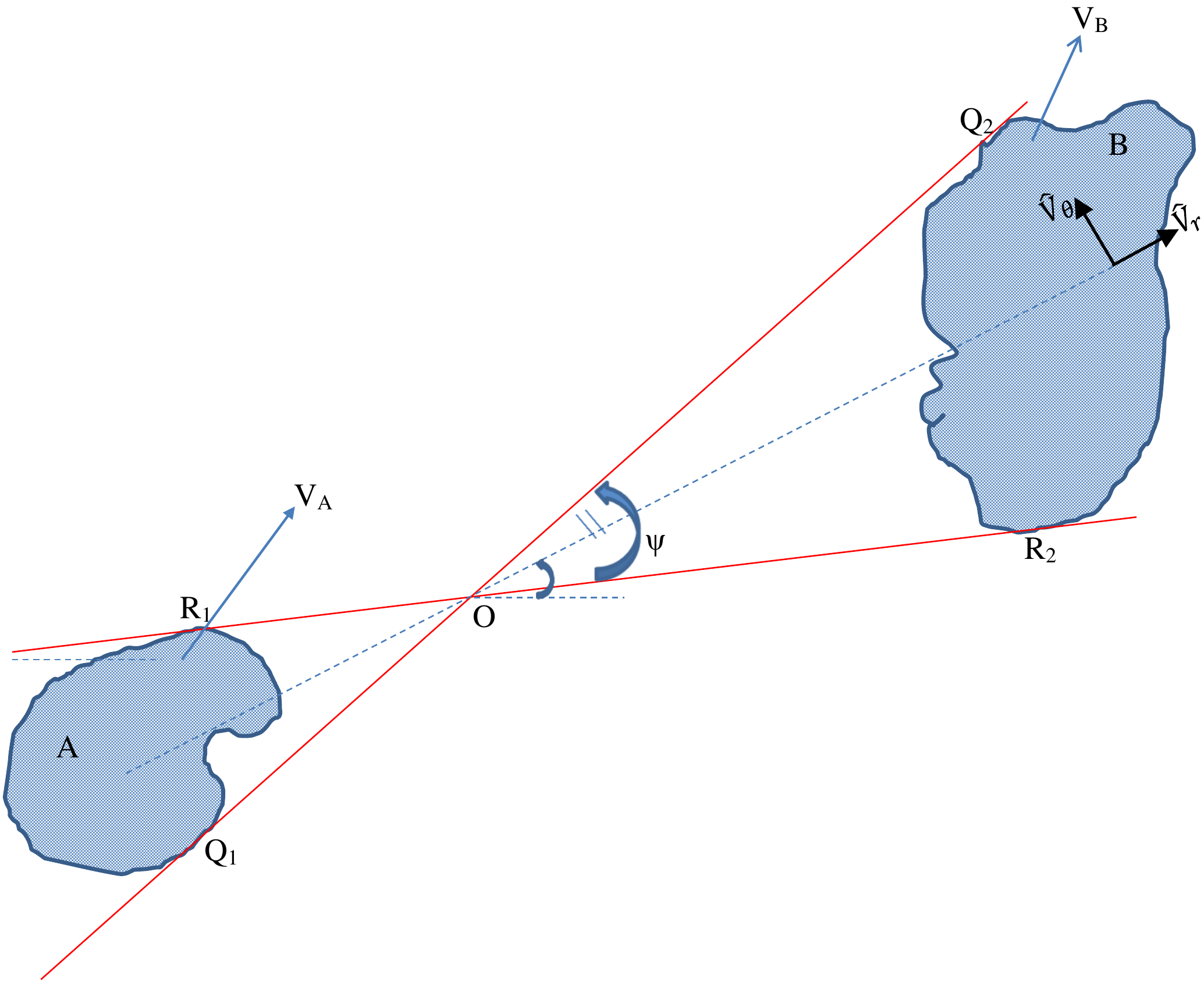}
\caption{Engagement geometry between arbitrarily shaped objects}
\label{fig:derivation}
\vspace{-15 pt}
\end{figure}

Refer Fig \ref{fig:derivation}, which shows two arbitrarily shaped objects $A$ and $B$ moving with velocities $V_A$ and $V_B$, respectively.  The lines $Q_1Q_2$ and $R_1R_2$ form a sector with the property that this represents the smallest sector that completely contains $A$ and $B$ such that $A$ and $B$ lie on opposite sides of the point of intersection $O$.  Let $\hat{V}_r$ and $\hat{V}_\theta$ represent the relative velocity components of the angular bisector of this sector and $V_r$ and $V_\theta$ represents the relative velocity components of line-of-sight, they are related as follows: %
\begin{equation}\label{eqn:VrVtheta}
    \begin{bmatrix}
    \hat{V}_r\\
    \hat{V}_\theta
    \end{bmatrix} = \begin{bmatrix}
    \cos\left({\theta}-\theta_b\right) & \sin\left({\theta}-\theta_b\right)\\
    -\sin\left({\theta}-\theta_b\right) & \cos\left({\theta}-\theta_b\right)
    \end{bmatrix}\begin{bmatrix}
    V_r\\
    V_\theta
    \end{bmatrix}
\end{equation}  As demonstrated in \cite{c2}, $A$ and $B$ are on a collision course if their relative velocities belong to a specific set.  This set is encapsulated in a quantity $y$ defined as follows:
\begin{equation}\label{eqn:y_hat}
    y = \dfrac{\hat{V}_\theta^2}{\left(\hat{V}_\theta^2 + \hat{V}_r^2\right)}-\sin^2\left({\dfrac{\psi}{2}}\right)
\end{equation}
The collision cone is defined as the region in the $(\hat{V}_\theta,\hat{V}_r)$ space for which $y < 0$, $\hat{V}_r < 0$ is satisfied. Thus, any relative velocity vector satisfying this condition lies inside the collision cone. The condition corresponding to $y=0$ and $\hat{V}_r<0$ defines the boundaries of the collision cone and any relative velocity vector satisfying this condition is aligned along the boundary of the cone. 

A challenge in computing the collision cone for arbitrarily shaped objects is in the computation  of the sector enclosing the objects $A$ and $B$ (shown in Fig \ref{fig:derivation}), and determination of the angle $\psi$. Note that as $A$ and $B$ move, the angle $\psi$ changes with time.  An iterative method to determine $\psi$ for arbitrarily shaped objects using the concept of conical hulls was presented in  \cite{tholen2018footprints}.  In \cite{ownCDCpaper},  an alternative, computationally light method to compute $\psi$ for objects modeled by 2-D quadric surfaces was presented. A method to compute 3-D collision cones for generic objects was presented in \cite{c4}.  In this paper, we extend the results from \cite{c2}, \cite{c4} and \cite{ownCDCpaper}, to present a  computationally inexpensive method to determine the 3-D collision cones corresponding to the objects discussed in Section \ref{eqn}.  

\vspace{-5pt}
\subsection{3D Collision Cone between two Ellipsoids}
We consider the scenario where $A$ and $B$ are  ellipsoids.  Let $\mathbf{Q_{eA}}$ and $\mathbf{Q_{eB}}$ represent the respective matrices corresponding to these ellipsoids.  A $3D$ collision cone between $A$ and $B$ can be generated by first computing the  $2D$ collision cones on several planes, and subsequently merging these $2D$ cones to get a combined cone in $3D$.  Without loss of generality, we stipulate that all planes contain the line joining the centers of $A$ and $B$, and each successive plane is  generated by rotating the preceding plane about this line. 

Let the centers of $A$ and $B$ be $(A_x,\ A_y,\ A_z)$ and $(B_x,\ B_y,\ B_z)$, respectively. Let $\mathbf{r}$ represent the vector joining these centers.  
Let $\mathbf{P_j}$ represent the $j^{th}$ plane, where $j \in \big\{1,2,\ldots,n \big\}$, and $n$ is the number of planes. 
%
\begin{figure}
\centering
\includegraphics[width=0.43\textwidth]{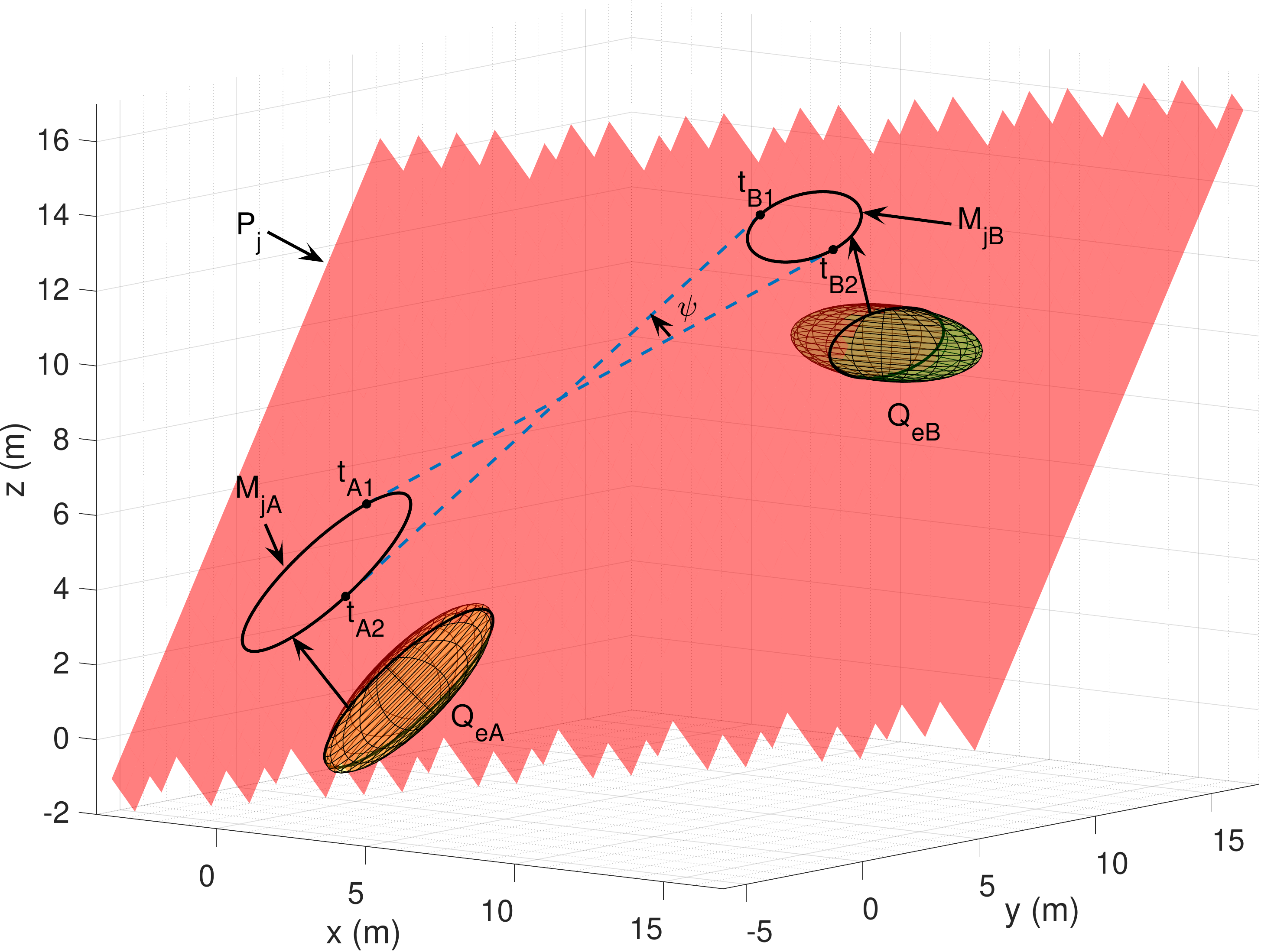}
\caption{Procedure to construct 3D collision cone between two ellipsoids}
\label{fig:planar_cross}
\vspace{-20pt}
\end{figure}
Refer Fig \ref{fig:planar_cross}, which shows one such plane. Let $\mathbf{R_{xj}}$ and $\mathbf{R_{yj}}$ be two mutually orthogonal unit vectors on this  plane. Here, we choose $\mathbf{R_{yj}} = \mathbf{r}$ and $\mathbf{R_{xj}}$ orthogonal to it, on that plane.  Define a matrix corresponding to the plane as follows:
$$
\mathbf{P_j} = \begin{bmatrix}
\mathbf{R_{xj}} & \mathbf{R_{yj}} & [A_x,\ A_y,\ A_z]^T \\
0 & 0 & 1
\end{bmatrix}
$$
The intersection of the plane $P_j$ with the two ellipsoids $A$ and $B$, will produce two ellipses.  The matrices $\mathbf{M_{jA}}$ and $\mathbf{M_{jB}}$ corresponding to these ellipses are found as follows:
\begin{eqnarray} \label{eqn:conic_sections}
   \mathbf{M_{jA}} = \mathbf{P_j}^T \mathbf{Q_{eA}} \mathbf{P_j},~~~ 
   \mathbf{M_{jB}} = \mathbf{P_j}^T \mathbf{Q_{eB}}\mathbf{P_j}
\end{eqnarray}
We then obtain the duals of these two ellipses as:
\begin{eqnarray} \label{eqn:conic_sections2}
   \mathbf{C_1}=\mathbf{M_{jA}}^{-1},~~~ \mathbf{C_2}=\mathbf{M_{jB}}^{-1}
\end{eqnarray}
Next, we compute the collision cone between $\mathbf{M_{jA}}$ and $\mathbf{M_{jB}}$.  For that, we compute the common tangents to these two ellipses, using algorithm \ref{algHM}.  This algorithm provides the points of tangency $t_{A1},t_{A2}$ that lie on $\mathbf{M_{jA}}$ and $t_{B1},t_{B2}$ that lie on $\mathbf{M_{jB}}$ (Please see Fig  \ref{fig:planar_cross} for an illustration).  
\begin{breakablealgorithm}
 		\caption{Common tangents to two ellipses}
	\begin{algorithmic}[1]
		\Statex // Given two matrices $\mathbf{C_1}$ and $\mathbf{C_2}$
			\Statex // Determine the degenerate conics
		\State $[U,\textbf{t}]  = eig(\mathbf{C_2}^{-1}  \mathbf{C_1}$) \Comment $\mathbf{U}=[\mathbf{u_1}\ \mathbf{u_2}\ \mathbf{u_3}],\ \mathbf{t} = [t_1,\ t_2,\ t_3]^T$
		\Statex // Perform Projective transformation of degenerate conic by $\mathbf{U}$
		\State $\mathbf{L}_i = \mathbf{U}^T (\mathbf{C_1} - \mathbf{t}_i \mathbf{C_2}) U$ for $i=1,2$
		\Statex // Find one of the intersection point of these projected degenerate conics
			\State $    x = \sqrt{-\dfrac{\mathbf{L}_2(3,3)}{\mathbf{L}_2(1,1)}},\ 
    y =  \sqrt{-\dfrac{\mathbf{L}_1(3,3)}{\mathbf{L}_1(2,2)}}$ 
    \Statex // Find all the tangent lines in projected coordinate system
   \State $    \mathbf{S} = \begin{bmatrix}
        x & y & 1\\
        -x & -y & 1\\
        -x & y & 1 \\
        x & -y & 1
        \end{bmatrix}$
    \Statex // Deproject the solution to homogeneous system
		\State $    \mathbf{S} = \mathbf{S} \mathbf{U}^T$
		\Statex // Define the centers of $\mathbf{M_{jA}}$ and $\mathbf{M_{jB}}$
	    \State	$    \mathbf{d_1} = \mathbf{C_1} [0\ 0\ 1]^T,\ \mathbf{d_1} = \mathbf{d_1}/\mathbf{d_1}(3)$  \State $\mathbf{d_2} = \mathbf{C_2} [0\ 0\ 1]^T,\ \mathbf{d_2} = \mathbf{d_2}/\mathbf{d_2}(3)$
	     \Statex // Determine which of the solutions are the inner tangents by using the property that two centers of the ellipses lie on opposite sides of the inner tangents. Defining a boolean vector
	     \State $bool = (\mathbf{Sd}_1)\odot (\mathbf{Sd}_2) < 0$ \Comment $\odot$ is Hadamard product
	   \Statex // Eliminating the indices of $s$ marked as false in $bool$ 
	     \State $\mathbf{S} =\mathbf{S}(bool)$ 
	      \Statex // Determine the homogeneous coordinates of tangent points on ellipses $\mathbf{M_{jA}}$ and $\mathbf{M_{jB}}$
	      \State $t_A = \mathbf{C_1 S}$, $t_B = \mathbf{C_2 S} $
 	\end{algorithmic}
 	\label{algHM}
\end{breakablealgorithm}
Note that $t_A$, $t_B$ $\in$ $\mathbf{R}^{3\times2}$ where each column denotes the homogeneous coordinates of each point of tangency. After obtaining these points of tangency, we obtain the tangent lines passing through these points and then find collision cone parameters $\psi$ and $\theta_b$ on that particular plane.  Henceforth, all the projected $2D$ states on a plane are marked by a subscript $P$.  Thus, $\psi_{Pj}$ and $\theta_{bPj}$ represent the values of $\psi$ and $\theta_b$ on plane $\mathbf{P_j}$.

\subsubsection*{Projection of $3D$ relative states into the $2D$ plane}
The final step of computing the $3D$ collision cone involves computing the projection of the $3D$ states of the objects $A$ and $B$ on each plane $P_j,j=\{1,\ldots,n\}$. Since by construction, all the planes contain the vector $\mathbf{r}$, so the length of the $3D$ vector $r$  is equal to its corresponding 2D projection, that is $r=r_p$. $\theta_p$ is LOS angle from $\mathbf{d_1}$ to $\mathbf{d_2}$.
Also, the relative velocity component along $\mathbf{r}$ in 3D (that is, $V_{r}$) will be equal to the relative velocity along LOS in 2D (that is, $V_{rp}$) and the $2D$ relative velocity perpendicular to $\mathbf{r}$, $V_{\theta p}$  acts along the direction of $\mathbf{R_{xj}}$.
Defining $V$ as the relative velocity vector between $A$ and $B$ as $\mathbf{V} = \mathbf{V}_B - \mathbf{V}_A$, $\hat{\mathbf{V}}_{\theta} = \mathbf{R_{xj}}$ as the unit vector along $V_{\theta2D}$, we get the following expressions for the projected $2D$ relative velocity components: 1) $V_{rp}  = V_r,\, 2) V_{\theta p} = (\mathbf{V}-{\mathbf{V}}_{r}) \cdot  \hat{\mathbf{V}}_{\theta}$
Then using (\ref{eqn:y_hat}), we can compute collision cone in $2D$. We repeat this process over all the $n$ planes to ultimately obtain $3D$ collision cone. 

\subsubsection*{Influence of $n$ on accuracy of 3D cone:} 
\begin{figure}
\centering
\includegraphics[width=0.45\textwidth]{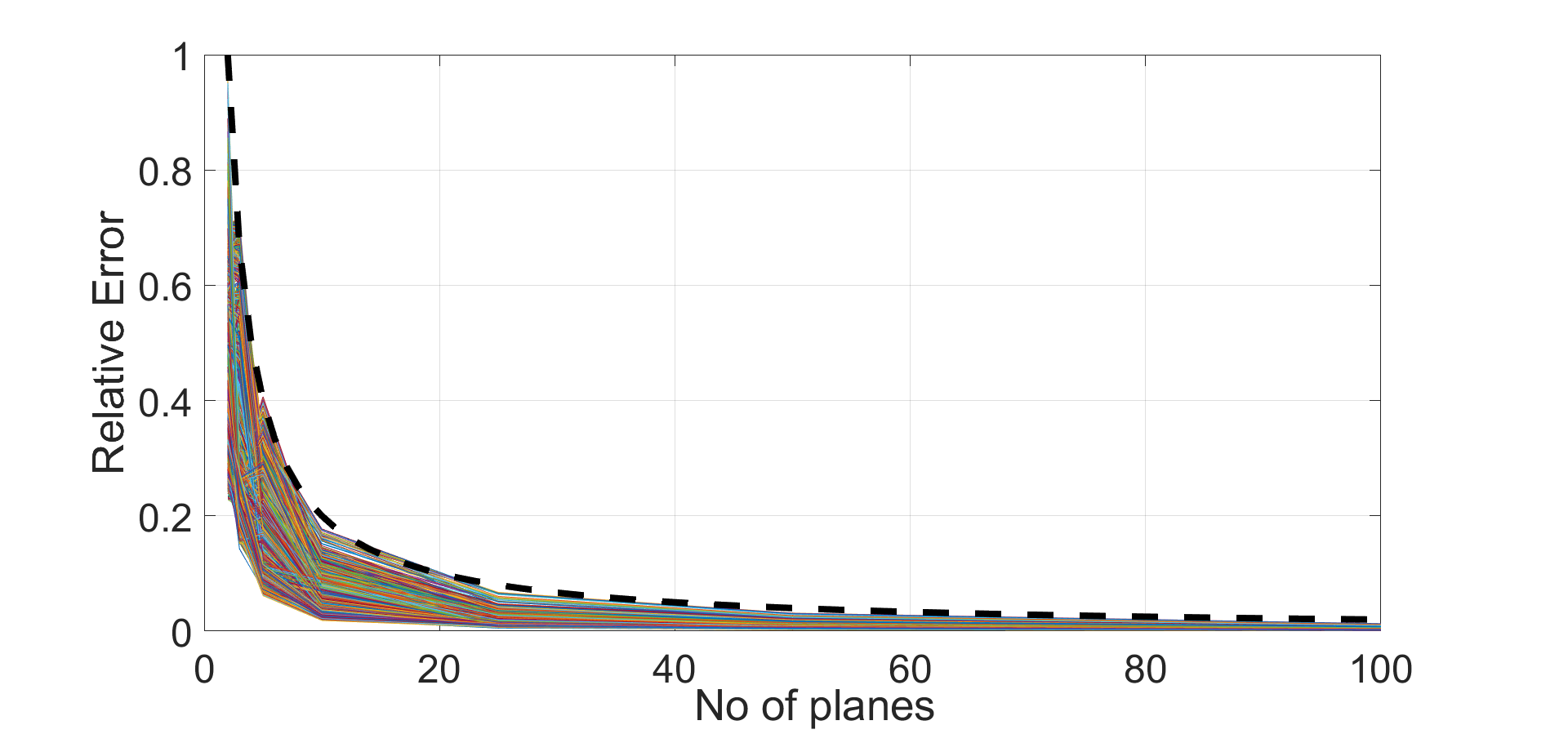}
\caption{Plot of relative error vs. no. of planes}
\label{fig:Plane_Error}
\vspace{-25pt}
\end{figure}

We note that by increasing the number of planes $n$, we can increase the accuracy of the collision cone, but this increases the computation time.  The proper choice of $n$ depends on the computational resources available and accuracy desired.  To evaluate the effect of $n$ on accuracy, a Monte Carlo simulation of $10,000$ different engagement geometries of two ellipsoids was performed, and for each engagement, the 3-D collision cone was computed for varying values of $n$. The cross-sectional area of the 3-D collision cone was computed in each case and this was used to determine the numerical accuracy as follows. The 3-D cone obtained with $n=360$ was treated as the truth model and the difference between the cross-sectional area of this cone, with the cone obtained using other values of $n$ is shown in Fig \ref{fig:Plane_Error} (the error is expressed as a  fraction). As seen in Fig \ref{fig:Plane_Error}, the error decreases rapidly with increasing $n$, and has an upper bound of $2/n$. This shows that a small number of planes can be used to compute the 3D collision cone with a relatively small error.

\subsection{3D Collision cone between an ellipsoid and a biconcave ellipsoid}\label{sec:ellip_biconellip}

\begin{figure}
\centering
\includegraphics[width=0.49\textwidth]{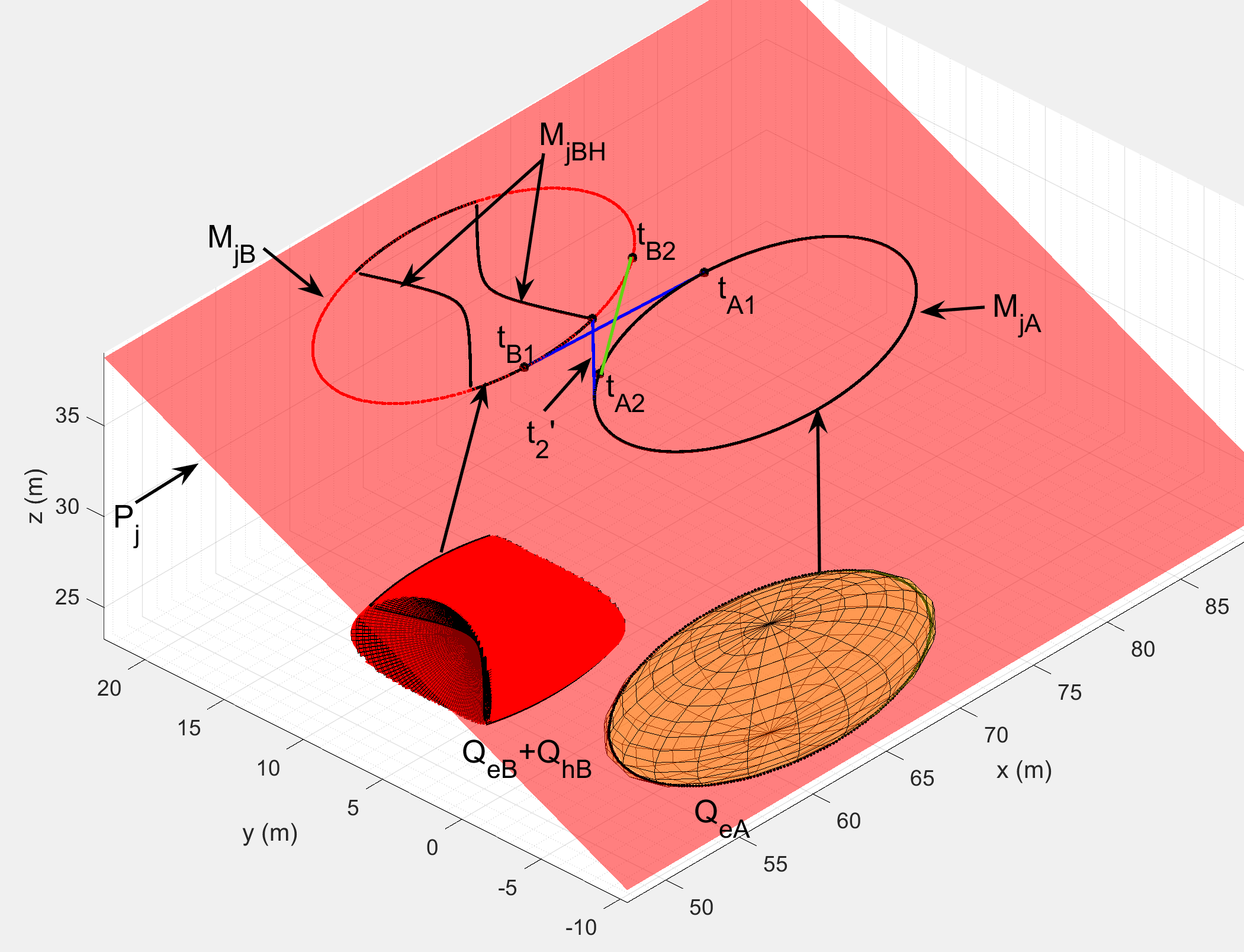}
\caption{3D collision cone between an ellipsoid and a biconcave ellipsoid}
\label{fig:planar_cross_confocal}
\vspace{-15pt}
\end{figure}

We next consider the scenario where $A$ is an  ellipsoid $A$ and $B$ is a biconcave ellipsoid.  Let the matrix corresponding to $A$ be $(\mathbf{Q_{eA}})$, and those corresponding to $B$ be $(\mathbf{Q_{eB}})$ and Hyperboloid $(\mathbf{Q_{hB}})$. To compute collision cone in this case, we take the planar cross-sections of the two objects similar to the case of two ellipsoids. First, we consider the engagement between $\mathbf{Q_{eA}}$ and $\mathbf{Q_{eB}}$ and define $C_1$ and $C_2$ on several planar cross-sections.  Refer Fig \ref{fig:planar_cross_confocal}, which shows the cross-sections on one such plane $\mathbf{P_j}$.  

We use  (\ref{eqn:conic_sections2}) and find the points of tangency on $\mathbf{M_{jA}}$ and $\mathbf{M_{jB}}$ using algorithm \ref{algHM}.  This will give us two points of tangency on each of $\mathbf{M_{jA}}$ (say $t_{A1}$ and $t_{A2}$) and $\mathbf{M_{jB}}$ (say $t_{B1}$ and $t_{B2}$).  We use these to define two \emph{candidate} tangent lines (say $t_1$ and $t_2$) passing through the pairs of points $(t_{A1},t_{B1})$ and $(t_{A2},t_{B2})$ respectively, and then check if $t_1$ and $t_2$ are valid tangent lines. 

Note that the planar cross-section of a biconcave ellipsoid can be either an ellipse or a combination of ellipse and hyperbola (say, a biconcave ellipse). If the planar cross section is an ellipse, then the candidate solutions $t_1$ and $t_2$ can be accepted as the valid solution on that plane.
However if that is not the case, then we check if the  points $t_{B1}$ and $t_{B2}$ lie on the biconcave ellipse. For this, we first determine the equation of the hyperbola on the plane $\mathbf{P_j}$ as:
\begin{eqnarray} \label{eqn:conic_sections_hyp}
  \mathbf{M_{jBH}} = \mathbf{P_j}^T \mathbf{Q_{eH}} \mathbf{P_j} 
\end{eqnarray}
and then check if 
$ \mathbf{t_{Bi}}^T \mathbf{M_{jBH}} \mathbf{t_{Bi}} < 0,\, i = 1,2 $

If either $t_{B1}$ and $t_{B2}$ satisfy the above equation then the corresponding tangent lines can be accepted as valid common tangents. The ones that do not satisfy the above equation need to be replaced by a new line tangent to $\mathbf{M_{jA}}$ passing through one of the corner points $(L,M,N,P)$ of the  biconcave ellipse. Note that the corner points can be obtained using algorithm \ref{algHM} by taking $\mathbf{C_1} = \mathbf{P_j^T Q_{eB} P_j}$ and $\mathbf{C_2} = \mathbf{P_j^T Q_{eH} P_j}$. After performing steps 1-5, we can obtain the desired corner points in the homogeneous system as $s$. To draw tangents from a corner point to $A$ we use a computationally efficient approach presented in~\cite{ownCDCpaper}. 
To proceed, we draw two tangents from each of the corner points to $\mathbf{M_{jA}}$. These tangent lines would be considered valid if: i) the centers of $\mathbf{M_{jA}}$ and $\mathbf{M_{jB}}$ lie on  opposite sides of each line, and ii) All the four corner lie on the same side of each line. We eliminate the candidates that do not satisfy these properties, and this will leave us with two inner common tangents from the four corner points to $\mathbf{M_{jA}}$.   Call these $t_1'$ and $t_2'$. If both $t_1$ and $t_2$ computed above are invalid then $t_1'$ and $t_2'$ can be accepted as the inner common tangents. However, if only one of them is invalid then that invalid tangent line should be replaced by $t_1'$ or $t_2'$, as the case may be. 
We finally use these computed tangents to calculate  $\psi_p$ and $\theta_{bp}$ in (\ref{eqn:y_hat}) to compute the collision cone.  A similar set of steps can be used to compute the collision cone between an ellipsoid and a biconvex hyperboloid.\\

\vspace{-10pt}
\subsection{3D Collision Cone between an ellipsoid and an arbitrarily shaped object}\label{sec:ellip_arbit}

\begin{figure}
\centering
\includegraphics[width=0.2\textwidth]{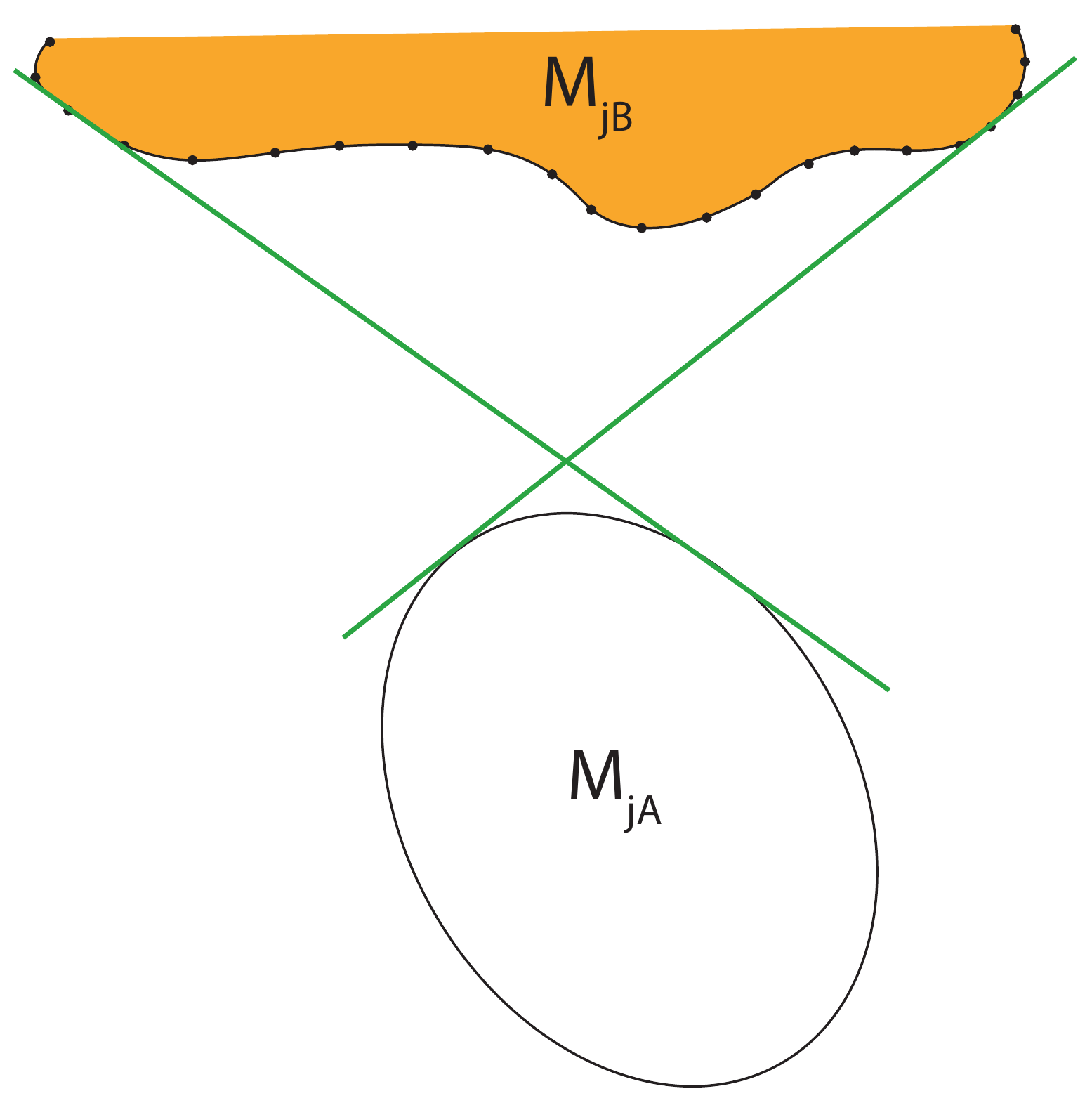}
\caption{Planar section of an ellipsoid and an arbitrary object}
\label{fig:arbitrary}
\vspace{-15pt}
\end{figure}

Let $A$ be an ellipsoid $(\mathbf{Q_{eA}})$ and $B$ represent an obstacle of arbitrary shape that cannot be represented  analytically. Assume that $A$ is equipped with multiple lidars and the point clouds from these lidars are fused using the Iterative Closest Point (ICP) algorithm~\cite{arun1987Least}. 
This 3D point cloud can then be used to determine the intersection of $B$ with a plane passing through the center of $A$ and $B$. Similar to the case of two ellipsoids, we obtain multiple planar cross-sections of $A$ and $B$. Planar cross-sections of $A$ corresponding to ellipse $\mathbf{M_{jA}}$ can be determined using (\ref{eqn:conic_sections}).  However  planar cross-sections of $B$ cannot be determined because it's shape is unknown and in fact, the only available knowledge of $B$ is of that  portion of $B$ which can be viewed by the lidars on $A$.  Let $\mathbf{M_{jB}}$ represent the portion of the planar cross section of $B$, which is visible from the lidars on $A$. A schematic is shown in Fig  \ref{fig:arbitrary}.

To obtain collision cone on that plane, we need to find the common tangents to $\mathbf{M_{jA}}$ and $\mathbf{M_{jB}}$. For this, we first draw four tangents, two from each of the extreme ends of $\mathbf{M_{jB}}$ to $\mathbf{M_{jA}}$. For each such line, we check if the centers of $\mathbf{M_{jA}}$ and $\mathbf{M_{jB}}$ lie on opposite sides of the line, and additionally, all points of $\mathbf{M_{jB}}$ lie on the same side. If these criteria are satisfied by any two lines, then these correspond to the desired common tangents and we can proceed towards computation of collision cone. Otherwise, we perform a search across the points of $\mathbf{M_{jB}}$ (starting from its extreme ends and moving towards the middle), and repeat this process until we find two common tangent lines.  We note that the points on the boundary of $\mathbf{M_{jB}}$ can be down-sampled to decrease the computation time. In most of the cases, the common tangents would pass through the points that are closer to the extreme ends of $\mathbf{M_{jB}}$, and so this algorithm would be able to find the solution in a few iterations. We point out that since this algorithm involves a search process, it will be computationally more expensive then the one used for quadrics in the preceding sections, but will still be computationally efficient.


\section{Collision Avoidance Acceleration} \label{guidance}
From (\ref{eqn:VrVtheta}) and (\ref{eqn:y_hat}),  we obtain the collision cone function $y_p$ for each plane so that it is now written in terms of the $2D$ states of the selected plane using the subscript $p$ as follows:
\begin{equation}\label{eqn:y}
        y_p =  \dfrac{\Big[ \splitfrac{ {V}_{\theta_p}^2 \cos^2({\theta_p}-\theta_{bp}) + V_r^2 \sin^2(\theta_p-\theta_{bp})  } {+2 V_r V_{\theta_p} \cos(\theta_p-\theta_{bp}) \sin(\theta_p-\theta_{pb})   }\Big]}{V_{rp}^2+V_{\theta p}^2}  -\sin^2\left({\dfrac{\psi_p}{2}}\right)
\end{equation}
The set of heading angles of $A$ satisfying $y_p<0,\hat{V}_{rp}<0$ on each plane are obtained, and then combined to get the $3D$ collision cone for that engagement.  If the velocity vector of $A$ lies inside the collision cone, then $A$ is on a collision course with $B$, and needs to apply a suitable lateral acceleration $a_{A}$ to drive steer its velocity vector out of the collision cone.  
We present a discussion on the choice of the avoidance plane (on which $a_{A}$ is to be applied), followed by analytical expressions for the lateral acceleration law required for collision avoidance.

\vspace{-5pt}
\subsection{Selection of Collision Avoidance Plane}

From the computed $3D$ collision cone, we take a $2D$ slice on a chosen plane, and then compute the lateral acceleration for avoidance on that plane.  We note that the choice of this plane can vary from one time instant to the next, and this is particularly true in dynamic environments and scenarios where the cross section of the 3D cone is not circular, or is non-convex (such as those discussed in this paper). There can be several ways to choose this plane. For instance, we can choose a plane on which the heading angle of the agent is closest to the boundary of the cone. Such a choice ensures that the angular deviation in the velocity vector of the vehicle (to get out of the cone) is small. In windy environments, the avoidance plane can be chosen such that the directions  of the applied lateral acceleration and the wind vector are not directly opposed to each other.  Other alternatives also exist.     

\subsection{Collision Avoidance Law}

On the chosen avoidance plane, $A$ needs to apply a suitable lateral acceleration $a_{A}$ to drive $y_p$ to a reference value $w \geq 0$, as this will be equivalent to steering its velocity vector out of the collision cone.  We employ dynamic inversion to determine this lateral acceleration.  Differentiating (\ref{eqn:y}),  we obtain the dynamic evolution of $y_p$ as follows:
\begin{equation}
\begin{split}
  \label{eq:yder}
    \dot{y}_p
 &=     \frac{\partial y_p}{\partial \theta_{bp}}\dot{\theta_{bp}} + \frac{\partial y_p}{\partial \theta_p}\dot{\theta_p} + \frac{\partial y_p}{\partial V_{\theta p}}\dot{V}_{\theta p}
 +\frac{\partial y_p}{\partial V_{rp}}\dot{V}_{rp} + \frac{\partial y_p}{\partial \psi_p}{\dot{\psi_p}}
\end{split}
\end{equation}
Define an error quantity $z_p(t) = y_p(t)-w$. Taking $w$ as a constant $\forall t$, we seek to determine $a_{lat,A}$ which will ensure the error $z_p(t)$ follows the dynamics $\dot{z}_p = -Kz_p$ where $K>0$ is a constant. This in turn causes the quantity $y$ to follow the dynamics $\dot{y}_p = -K(y_p-w)$.
%
Note that all the partial derivatives of $y_p$ can be computed analytically from (\ref{eqn:y}).  While the state kinematic equations are given in (\ref{eq:state_kinematics}), we however do not have analytical expressions of $\dot{\theta}_{bp}$ and $\dot{\psi}_p$ and these will have to be synthesized numerically. 

\underline{Non-cooperative Collision Avoidance:}
Here, the onus is on $A$ to apply a lateral acceleration to steer its velocity vector out of the collision cone, and $B$ does not cooperate.  Substituting partial derivatives and state derivatives in (\ref{eq:yder}) and assuming $a_B=0$, we get an expression for $a_{A}$ as:
\begin{equation}\label{eqn:guidance_law}
a_{A} = -(V_{rp}^2+V_{\theta p}^2) \frac{N_1 + N_2}{D_1 D_2}
\end{equation}
where, the quantities $N_1$, $N_2$, $D_1$, $D_2$ are as follows:
\begin{equation}
\begin{split}
  N_1 &= (V_{rp}^2+V_{\theta p}^2) (2K(w-y_p) + \dot{\psi_p} \sin(\psi_p)),\, N_2 =  2 \dot{\theta_b}_p D_1  \nonumber \\
 D_1 &= 2 V_{rp} V_{\theta p} \cos(2(\theta_p-\theta_{bp}))
 + (V_{rp}^2-V_{\theta p}^2) \sin(2(\theta_p-\theta_{bp})) \nonumber\\
  D_2 &= 2 \left( V_{rp} \cos(\alpha_{pA}-\theta_p) + V_{\theta p} \sin(\alpha_{pA}-\theta_p)   \right) \label{D2}
  \end{split}
\end{equation}

\underline{Cooperative Collision Avoidance:} Here, $A$ and $B$ cooperate with one other in applying suitable lateral accelerations so that they jointly steer their velocity vectors out of the collision cone.  Assume that $\mu$ represents the acceleration ratio, that is, $\mu=a_B/a_A$.    
Substituting partial derivatives and state derivatives in (\ref{eq:yder}), we get equations for $a_{A}$, $a_{B}$ as:
\begin{equation} \label{eqn:guidance_law2}
\begin{split}
    &0.5 a_{A}\ D_{1} + a_{B}\ N_3 = -\frac{N_1+\dot{\theta}_{bp} D_1 (V_{rp}^2+V_{\theta p}^2)}{D_1},\text{where}\\
    &N_3 = -(V_{rp} \cos (\alpha_{pB} -\theta_p )-V_{\theta p} \sin (\alpha_{pB} -{\theta p} ))
\end{split}
\end{equation} 
Using the acceleration ratio $\mu$, the above leads to the following accelerations:
\begin{equation}\label{eqn:guidance_law2b}
\begin{split}
    a_{A} = -\frac{N_1+\dot{\theta}_{bp} D_1 (V_{rp}^2+V_{\theta p}^2)}{D_1(0.5 D_1+\mu N_3)} ,\ 
    a_{B} = \mu a_A
\end{split}
\end{equation} 

\underline{Computation of Direction of Acceleration Vectors :} 
These computed $a_{A}$ and $a_B$ are applied at angles of $(\alpha_{pA}+\pi/2)$ and $(\alpha_{pB}+\pi/2)$, respectively, on the selected plane.  Here, $\alpha_{pA}$ and $\alpha_{pB}$ represent the heading angles of $A$ and $B$ on that plane.  In $3D$, the  direction of the applied acceleration is as follows:
\begin{eqnarray}
\hat{a}_{i} &=& \mathbf{P_j}\ [\cos(\pi/2+\alpha_{pi})\ \sin(\pi/2+\alpha_{pi})\ 0]^T \nonumber \\
\delta_{i} &=& \tan^{-1}\left(\frac{\hat{a}_{i}(2)}{\hat{a}_{i}(1)}\right),\, 
\gamma_{i} = \sin^{-1}\left(\frac{\hat{a}_{i}(3)}{\Vert\hat{a}_{i}\Vert} \right),\, i=A,B  \nonumber 
\end{eqnarray}

 \begin{figure*}
\centering
\includegraphics[width=0.8\textwidth]{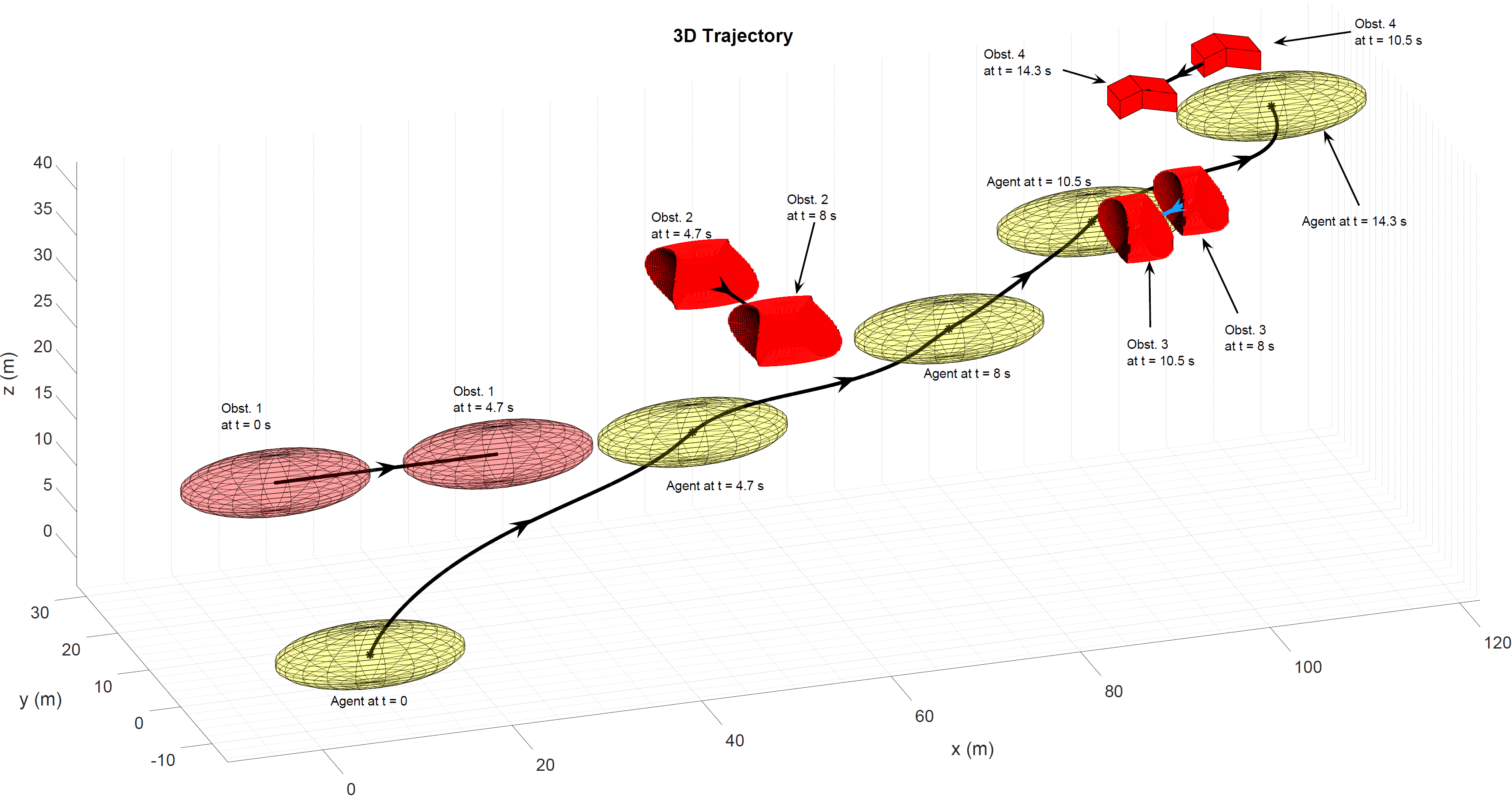}
\caption{Simulation 1: 3D Trajectory}
\label{fig:trajec}
\end{figure*}
\vspace{-5pt}
\section{Simulation Results}\label{sims}
\begin{figure}
\centering
\includegraphics[width=0.35\textwidth]{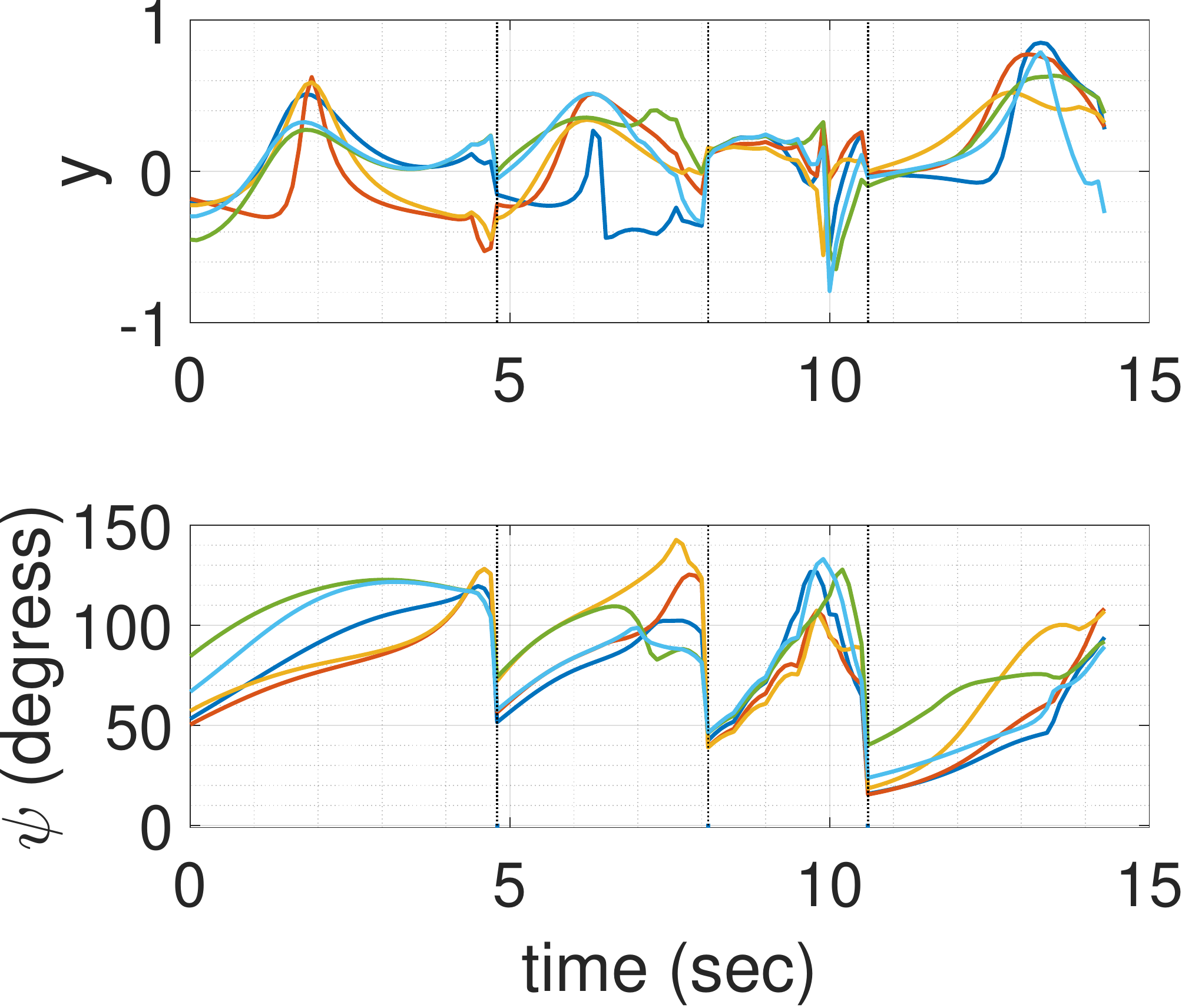}
\caption{Simulation 1: Collision cone $y$ and $\psi$ on multiple planes}
\label{fig:y_psi_multi}
\vspace{-20pt}
\end{figure}

\begin{figure}
\centering
\includegraphics[width=0.35\textwidth]{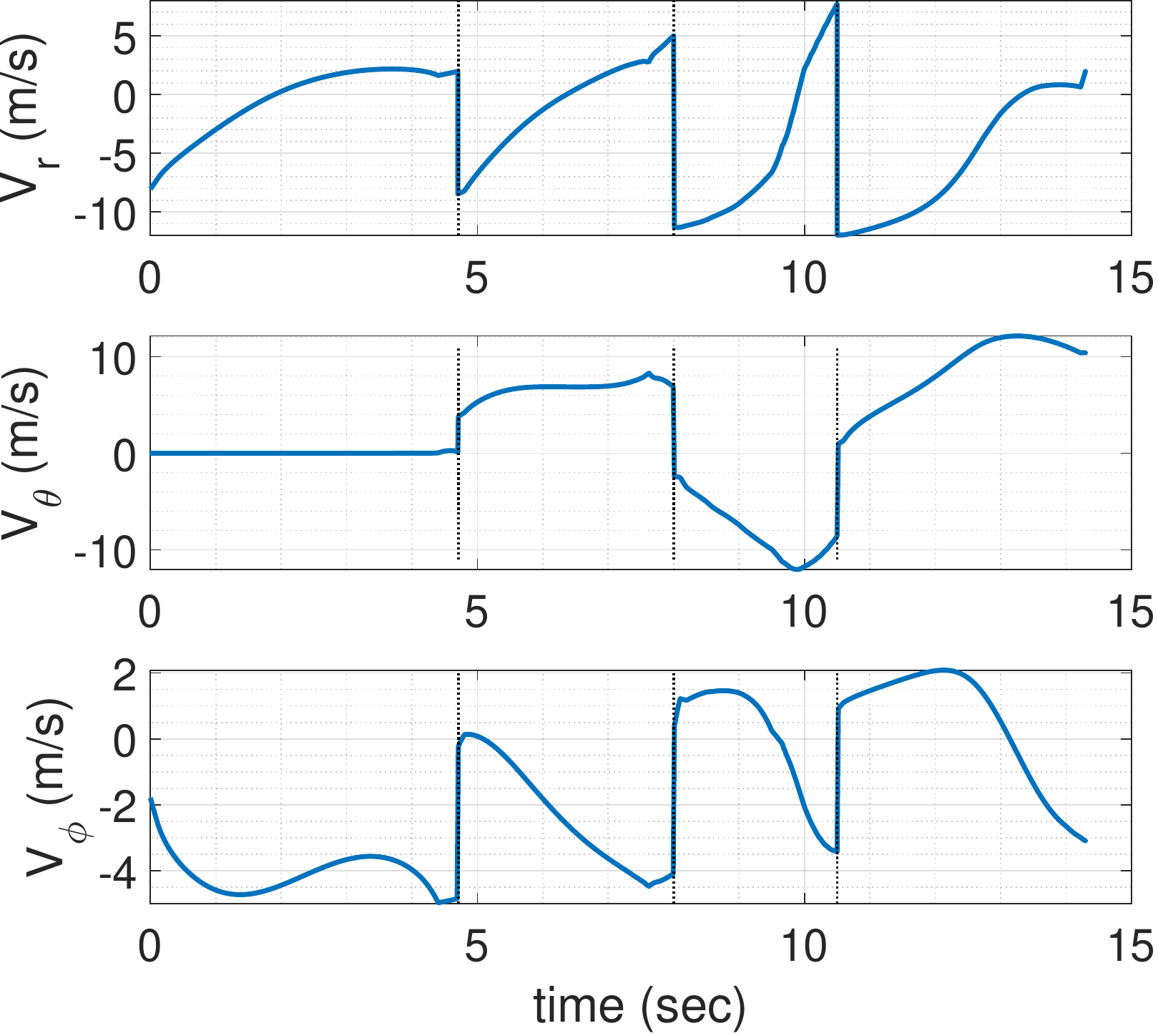}
\caption{Simulation 1: Time histories of $V_r$, $V_\theta$ and $V_\phi$}
\label{fig:Vr_Vtheta_Vphi}
\vspace{-10pt}
\end{figure}

\begin{figure}
\centering
\includegraphics[width=0.35\textwidth]{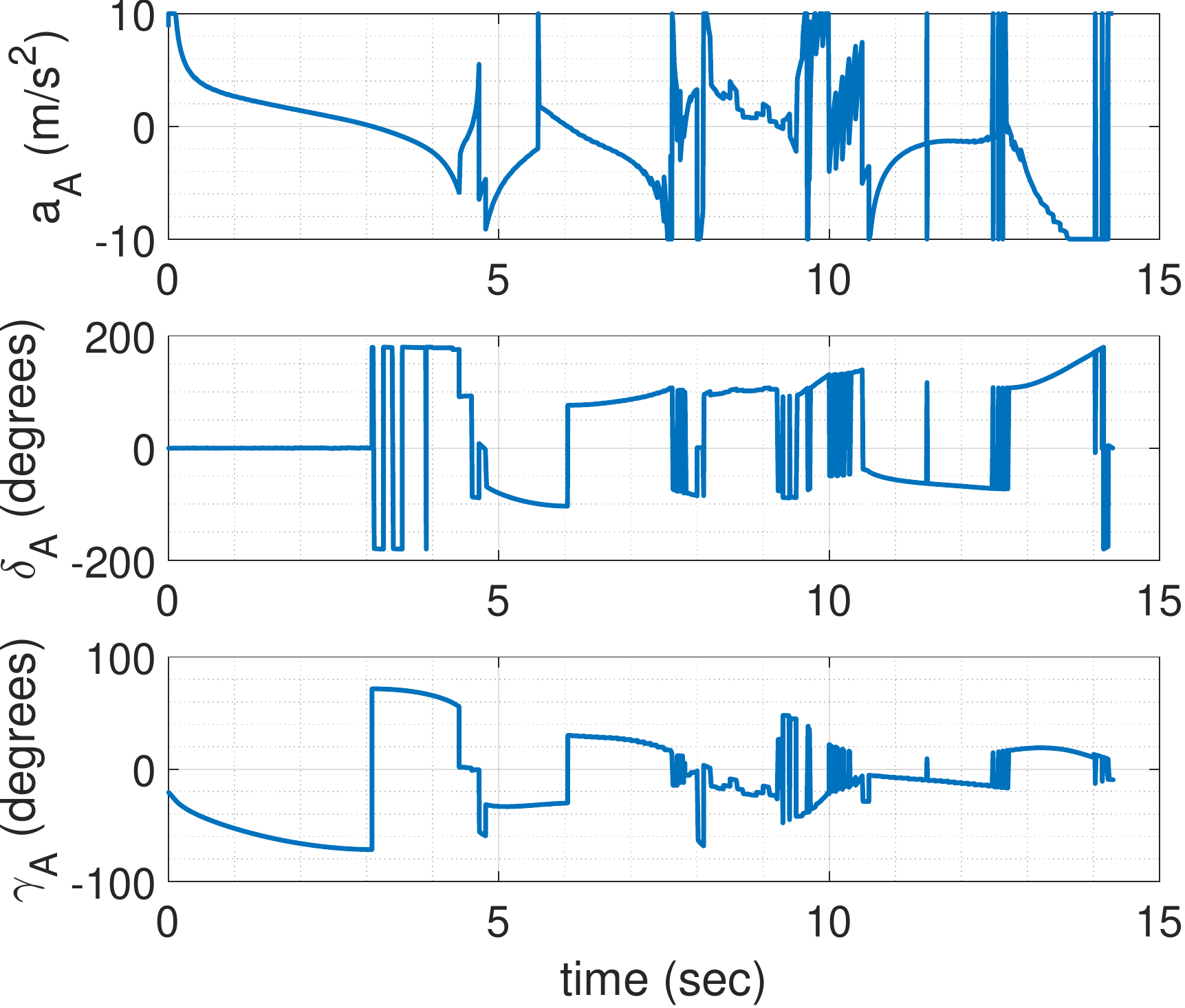}
\caption{Time histories of acceleration and it's direction}
\label{fig:acc_3D}
\vspace{-20pt}
\end{figure}

\begin{figure}
\centering
\includegraphics[width=0.35\textwidth]{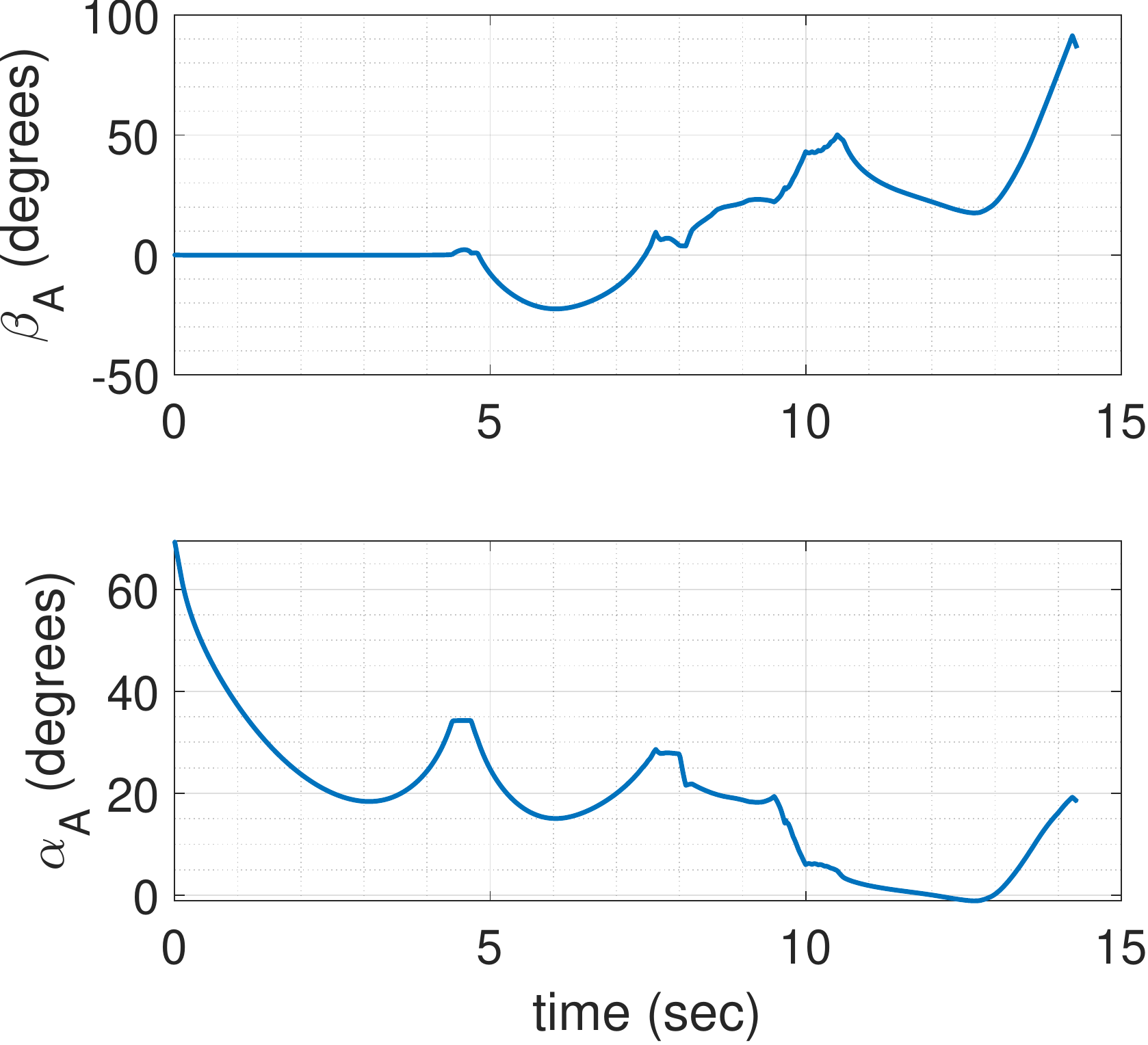}
\caption{Simulation 1: Time histories of azimuth and elevation of heading}
\label{fig:alpha_beta}
\vspace{-10pt}
\end{figure}

\begin{figure}
\centering
\includegraphics[width=0.35\textwidth]{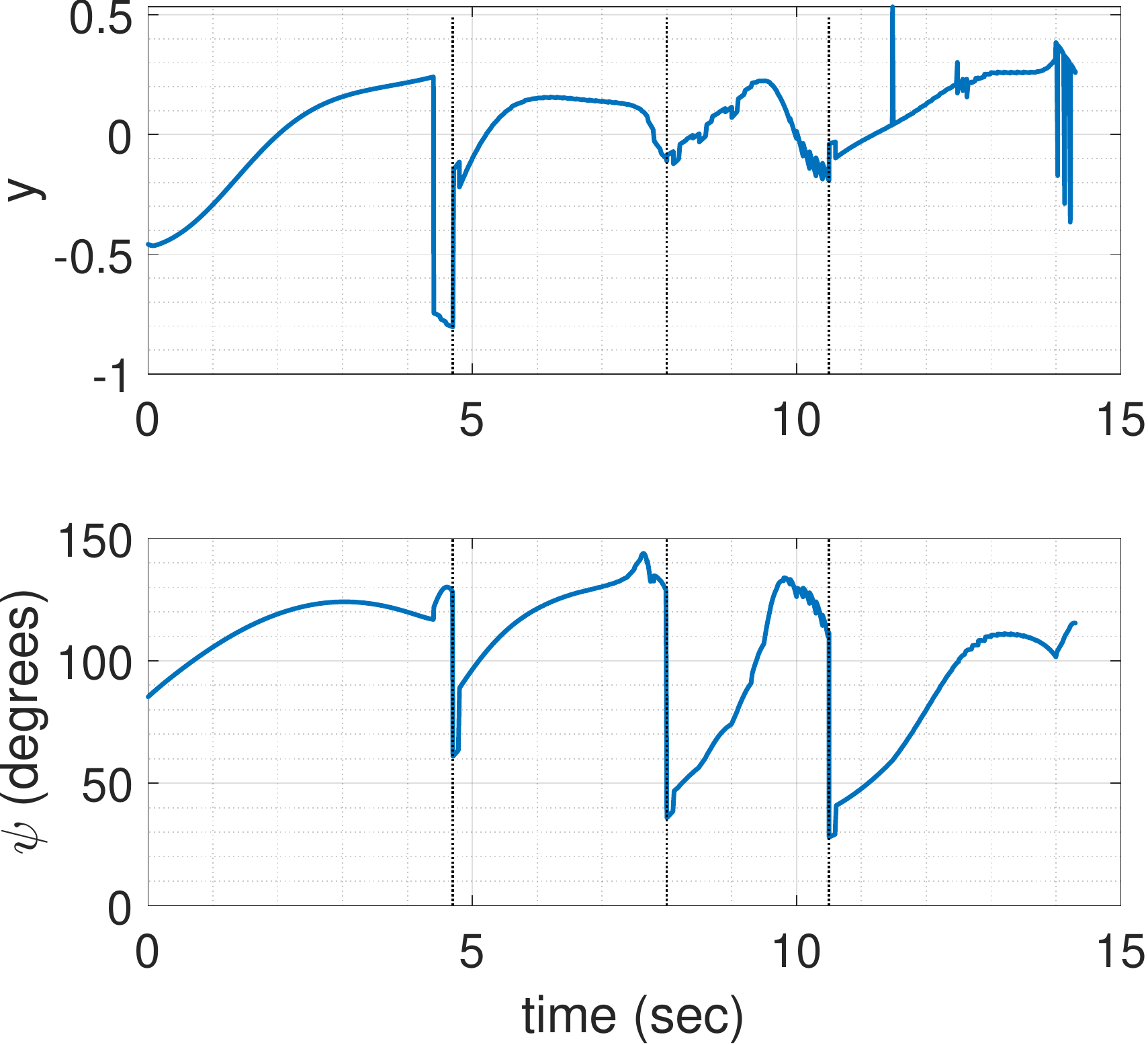}
\caption{Simulation 1: Time histories of $y$ and $\psi$ on maximum $\psi$ plane}
\label{fig:y_psi_max_plane}
\vspace{-10pt}
\end{figure}

Two simulation cases for a  non-cooperative and cooperative collision avoidance scenario, respectively, are presented. In the non-cooperative case, an engagement geometry is chosen where the agent $A$ is an ellipsoid with center at $(10,0,0)$ at initial time $t=0$ and semi-principal axes of $10\, m,\, 5\, m,\, 3\, m$. Its speed is $8.5 \,\,\mathrm{m/s}$ with the initial heading direction at an azimuth of $0^\circ$ and elevation of $69^\circ$, as seen in Fig~\ref{fig:alpha_beta}. $A$ faces a series of four obstacles $B,C,D$ and $E$ in quick succession.  $B$ is an ellipsoid with it's center initially located at $(0, 0, 20)$ and having the same principal axes as $A$. Its speed is $5\,\, \mathrm{m/s}$ with an initial heading angle at an azimuth and elevation of $0^\circ$. For this engagement, the collision cone $y_p$ is computed on multiple planes using the algorithms of Section~\ref{sec:Collcone} as shown in Fig.~\ref{fig:y_psi_multi}. It can be clearly seen that the value of $y_p$ obtained from those planes is negative at $t=0$. Since $V_r$ is also negative at $t=0$ (See Fig~\ref{fig:Vr_Vtheta_Vphi}), both ellipsoids are on a collision course.  In order to avoid $B$, the plane with the maximum value of $\psi_p$ (at each instant) is selected as the avoidance plane.  The collision cone $y_p$ along with the projected value of the states is used to generate the acceleration command using (\ref{eqn:guidance_law}) and the time history of the magnitude and direction of this acceleration is given Fig~\ref{fig:acc_3D}. Fig.~\ref{fig:y_psi_max_plane}  shows the time history of $y_p$ computed on the plane of maximum $\psi_p$.  From this, and the plot of $V_r$ in Fig~\ref{fig:Vr_Vtheta_Vphi}, it can be seen that both parameters become greater than zero after a certain time, signifying a successful collision avoidance maneuver.  $A$ is able to fully steer away from $B$ after $4.7\,\mathrm{s}$. The next obstacle $C$ is a biconcave ellipsoid with center at $(50,15,30)$ at time $t = 4.7\,\mathrm{s}$. It's speed is $5.1\,\mathrm{m/s}$ with heading angle at an azimuth of $0^\circ$ and elevation of $34.5^\circ$. Again from the $y_p$ plot in Fig~\ref{fig:y_psi_multi}, it can be seen that at $t = 4.7\,\mathrm{s}$, $A$ is on a collision course with $C$. The collision cone $y_p$ and value of $\psi_p$ on various planes is obtained using the steps described in Section \ref{sec:ellip_biconellip}. The avoidance  acceleration is again computed on the plane of maximum $\psi_p$  and it can be seen that the heading direction vector is steered out of the cone. At $t = 8\,\mathrm{s}$, $A$ encounters the next obstacle $D$ which is a shape-changing biconcave ellipsoid, whose shape transitions from an ellipsoid to a biconcave ellipsoid with varying levels of concavity.  It's center is at $(95,-5,40)$ and speed is $3.61\,\mathrm{m/s}$ with heading direction at an azimuth of $5^\circ$ and an elevation of $28^\circ$. Similar to the first two cases, avoidance acceleration  is computed on the plane of maximum $\psi_p$ to steer the velocity vector of $A$ out of the collision cone. After $10.5\,\mathrm{s}$, the agent encounters the fourth obstacle $E$ which is a $10$ faced polyhedron with center at $(110,30,40)$ and having the speed and direction as $C$. Collision cone $y_p$ and $\psi_p$ on the various planes is obtained using the steps outlined in Section ~\ref{sec:ellip_arbit}. Acceleration is applied to avoid the obstacle and the obstacle is successfully cleared. The trajectories of the agent $A$ and all the obstacles can be viewed in Fig~\ref{fig:trajec}.

In the Cooperative Collision Avoidance case, $A$ is an ellipsoid, and encounters two other agents $B_1$ (an ellipsoid) and $B_2$ (a biconcave ellipsoid) in quick succession. Both pairs of agents apply avoidance accelerations cooperatively using  (\ref{eqn:guidance_law2b}). 
Due to page constraints, the plots for this case are not given in the paper, but are shown in the accompanying video. 

\section{CONCLUSIONS} \label{conclusions}

In this paper, we model 3-dimensional objects having  elongated and/or non-convex shapes, by using an appropriate combination of ellipsoids and one-sheeted or two-sheeted hyperboloids.  The use of ellipsoids and hyperboloids  provides a much tighter and less-conservative approximation to the shapes of such objects.  This increases the amount of free space available for the robot trajectories.  We demonstrate the construction of 3-D collision cones for such objects and present collision avoidance laws, for both cooperative as well as non-cooperative collision avoidance.  Simulation results are presented.  

\vspace{-10pt}









\bibliography{IEEEexample}
\bibliographystyle{IEEEtran}

\end{document}